\documentclass{article} 
\usepackage{iclr2026_conference,times}

\usepackage{amsmath,amsfonts,bm}

\def\eqref#1{equation~\ref{#1}}

\def\1{\bm{1}}

\DeclareMathAlphabet{\mathsfit}{\encodingdefault}{\sfdefault}{m}{sl}
\SetMathAlphabet{\mathsfit}{bold}{\encodingdefault}{\sfdefault}{bx}{n}

\usepackage{hyperref}
\usepackage{url}
\usepackage{graphicx}
\usepackage{algorithm}
\usepackage{algpseudocode}

\title{Self-Supervised JEPA-based World Models for LiDAR Occupancy Completion and Forecasting}

\author{Haoran Zhu, Anna Choromanska\\
Department of Electrical and Computer Engineering
\\
New York University\\
\texttt{\{hz1922,ac5455\}@nyu.edu} \\
}

\iclrfinalcopy 
\begin{document}

\maketitle

\begin{abstract}
Autonomous driving, as an agent operating in the physical world, requires the fundamental capability to build \textit{world models} that capture how the environment evolves spatiotemporally in order to support long-term planning. At the same time, scalability demands learning such models in a self-supervised manner; \textit{joint-embedding predictive architecture (JEPA)} enables learning world models via leveraging large volumes of unlabeled data without relying on expensive human annotations. In this paper, we propose \textbf{AD-LiST-JEPA}, a self-supervised world model for autonomous driving that predicts future spatiotemporal evolution from LiDAR data using a JEPA framework. We evaluate the quality of the learned representations through a downstream LiDAR-based occupancy completion and forecasting (OCF) task, which jointly assesses perception and prediction. Proof of concept experiments show better OCF performance with pretrained encoder after JEPA-based world model learning.
\end{abstract}

\section{Introduction}
World models are becoming an increasingly important direction for autonomous driving, as they enable the prediction of spatial and temporal scenes, allowing vehicles to drive in imaginary space and simulate the future. Currently, most driving world models~\cite{hu2023gaia, russell2025gaia} are generative world models that explicitly generate pixel-level details for predicted future frames. However, this approach is computationally expensive to train and may suffer from hallucinations, producing physically implausible and inconsistent videos. On the other hand, some world models predict future states in latent space instead~\cite{li2024enhancing, zheng2025world4drive}, which shows promise in terms of computational efficiency while ignoring certain pixel-level details.

However, representation collapse is a challenge in training latent world models, where all representations produced by the encoder collapse to constant vectors. LAW~\cite{li2024enhancing} mitigates this issue by introducing an additional future waypoint regression loss, supervised by human drivers’ waypoints, alongside the latent prediction loss. Such waypoints are often noisy, may be suboptimal, and in real-world scenarios may admit multiple plausible futures. Therefore, this training strategy may limit the expressiveness of the encoder’s representations. In contrast, World4Drive~\cite{zheng2025world4drive} avoids representation collapse by utilizing auxiliary pretrained encoders, which require labeled data for training and thus limit scalability.

In this paper, we present \textbf{AD-LiST-JEPA} (\textbf{A}utonomous \textbf{D}riving with \textbf{Li}DAR data in a \textbf{S}patio\textbf{T}emporal setting via \textbf{J}oint \textbf{E}mbedding \textbf{P}redictive \textbf{A}rchitecture~\cite{lecun2022path}). JEPA is a paradigm that learns world models entirely in latent space without representation collapse, either by using a moving-average update of the target encoder or by maximizing the information captured by the encoder through explicit regularization, such as variance regularization. Previous work, AD-L-JEPA~\cite{zhu2025ad}, explores JEPA for self-supervised representation learning in an autonomous driving single-frame LiDAR setting. AD-LiST-JEPA pushes this further by learning a latent world model that predicts future spatiotemporal representations from multi-frame LiDAR data.


\section{Related Works}
\vspace{-0.1in}
\subsection{JEPA-based World Models}
World Models aim to predict the spatiotemporal evolution of an environment. At a high level, world models can be classified into \emph{generative world models}~\cite{bruce2024genie, hu2023gaia}, which generate pixel-level details of predictions, and \emph{non-generative world models}, which predict future representations at the latent level, e.g, \emph{joint-embedding predictive architectures (JEPA)}~\cite{lecun2022path}. World models based on JEPA propose predicting embeddings of masked or future image patches in a self-supervised manner~\cite{bardes2024revisiting} to capture multiple possible future states, rather than explicitly predicting data in pixel space. Recently, JEPA has been applied across different modalities, such as images~\cite{assran2023self} and videos~\cite{bardes2024revisiting,assran2025v}. In the context of autonomous driving,\cite{zhu2025ad} explores LiDAR-based data in single-frame settings and evaluates the representation quality on a downstream LiDAR-based 3D object detection task. JEPA has shown promising results in leveraging world models for planning~\cite{zhou2024dino, sobal2025learning}.

\subsection{LiDAR-based Future Forecasting Tasks}
LiDAR is a key sensor in autonomous driving. Unlike camera images, LiDAR data are represented as sparse point clouds. To evaluate LiDAR-based world models, common downstream tasks include point cloud forecasting, occupancy forecasting, and occupancy completion (OCF). Point cloud forecasting directly predicts future point clouds, but as point clouds are strongly correlated with LiDAR sampling patterns (ray casting) and sensor intrinsics~\cite{khurana2023point}. Occupancy forecasting~\cite{khurana2023point, agro2024uno} mitigates this issue by predicting occupancy in a fixed-range bird’s-eye-view (BEV) grid or 3D voxel space, where a cell is considered occupied if it contains at least one point. Due to sparsity and occlusions, occupancy forecasting captures only partial object geometry. OCF~\cite{liu2024lidar} extends occupancy forecasting by temporally aggregating points from the same instance over a long horizon, producing more complete occupancy labels. The downstream task is to predict this completed occupancy over time. As OCF evaluates both perception and prediction while providing more reliable scene geometry, we adopt it as a downstream task to assess world model capability in this paper.

\section{Method}
We propose a simple framework for self-supervised JEPA world model pretraining and supervised downstream OCF fine-tuning to evaluate the capability of world models. The framework consists of two phases: (1) self-supervised world modeling and (2) supervised downstream OCF finetuning. After Phase 1, we evaluate the JEPA-based world models by reusing the pretrained encoder for Phase 2 fine-tuning.

\subsection{Phase 1: Self-Supervised World Modeling}
The design philosophy is to train the network to predict the embeddings of masked multi-frame point cloud regions given the unmasked multi-frame point cloud regions, in order to capture the spatio-temporal evolution of the environment. To enable learning better representations, the novelty of our method lies in proposing a group BEV-guided masking strategy across frames that separates the ego vehicle from the world for JEPA training.

\subsubsection{Group BEV-Guided Masking}
For the masking, on the one hand, the data collected at each time frame consist of ego-centric point clouds, which reflect both the evolution of the environment and the transformation of the ego vehicle from one frame to the next. These ego-motion transformations are recorded and known by the autonomous driving system~\cite{khurana2023point}. Therefore, to learn better representations, the ego vehicle should be separated from the  world~\cite{sobal2022separating}. On the other hand, if we directly extend AD-L-JEPA's single-frame modified BEV-guided masking~\cite{zhu2025ad, lin2022bev}, which creates masks at BEV grids for both empty and non-empty cells, to multi-frame settings by propagating masks at the same spatial grid across time in ego-centric point clouds, the motion of the ego vehicle will cause information leakage across time steps. To address this issue, we propose transforming each point \( p_j \) in the multi-frame data from time \( -T \) to \( T \) into a common coordinate system at time \( 0 \) using the transformation
$
p_j' \leftarrow R^{\mathsf{T}} p_j + c,
$
where \( R \) is the rotation matrix and \( c \) is the translation matrix.

\begin{figure}[h]
\label{fig:group_masking}
\begin{center}
\includegraphics[width=\linewidth]{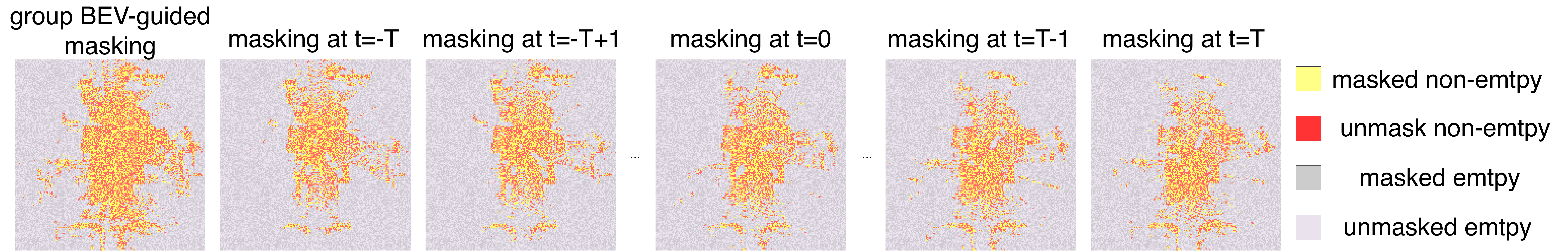}
\end{center}
\caption{Proposed group BEV-guided masking.}
\label{fig:proposed_group_bev_guided_masking}
\end{figure}

After the coordinate transformation, another challenge in multi-frame LiDAR data is that the number of non-empty grids (i.e., grids that contain points) may vary across frames, even for the same spatial location. Therefore, we propose a group masking strategy across all frames. Specifically, we first map all points from all frames to spatial grids to determine whether each grid is empty or non-empty at the group level. We then propagate this group masking to each individual frame. For any grid that is non-empty in the group masking (i.e., contains an object across time) but is empty (caused by occlusion) in a particular frame, the masking state of that grid in that frame is determined by the group masking, while the grid itself is still treated as empty in that frame. The visualization of the proposed group masking is in Figure~\ref{fig:proposed_group_bev_guided_masking}.

\subsubsection{Netowrk Architecture}
The architecture used in Phase 1 is inherited and adapted from AD-L-JEPA~\cite{zhu2025ad}, which is designed for single-frame JEPA-based LiDAR representation learning. Notably, the context encoder and the backbone of the target encoder are identical to those of the single-frame sparse 3D convolutional network used in AD-L-JEPA. This design choice is intentional because, in autonomous driving settings, computational resources are limited. As a result, most existing state-of-the-art architectures~\cite{hu2023planning, li2024hydra} employ single-frame encoders rather than multi-frame encoders and aggregate features after independently encoding each frame. The key difference in our approach is that the input point clouds are multi-frame rather than single-frame. Masking is performed at the BEV grid level using our proposed group BEV-guided masking strategy and then upsampled back to the original point cloud to form multi-frame context and target point clouds. We reshape the output 3D voxel representations of each frame into per-frame BEV representations and concatenate them along the height dimension to obtain multi-frame BEV representations. Finally, a simple 3D convolutional predictor is applied to predict the target multi-frame BEV representations.

\begin{figure}[ht!]
\vspace{-0.5in}
\begin{center}
\includegraphics[width=\linewidth]{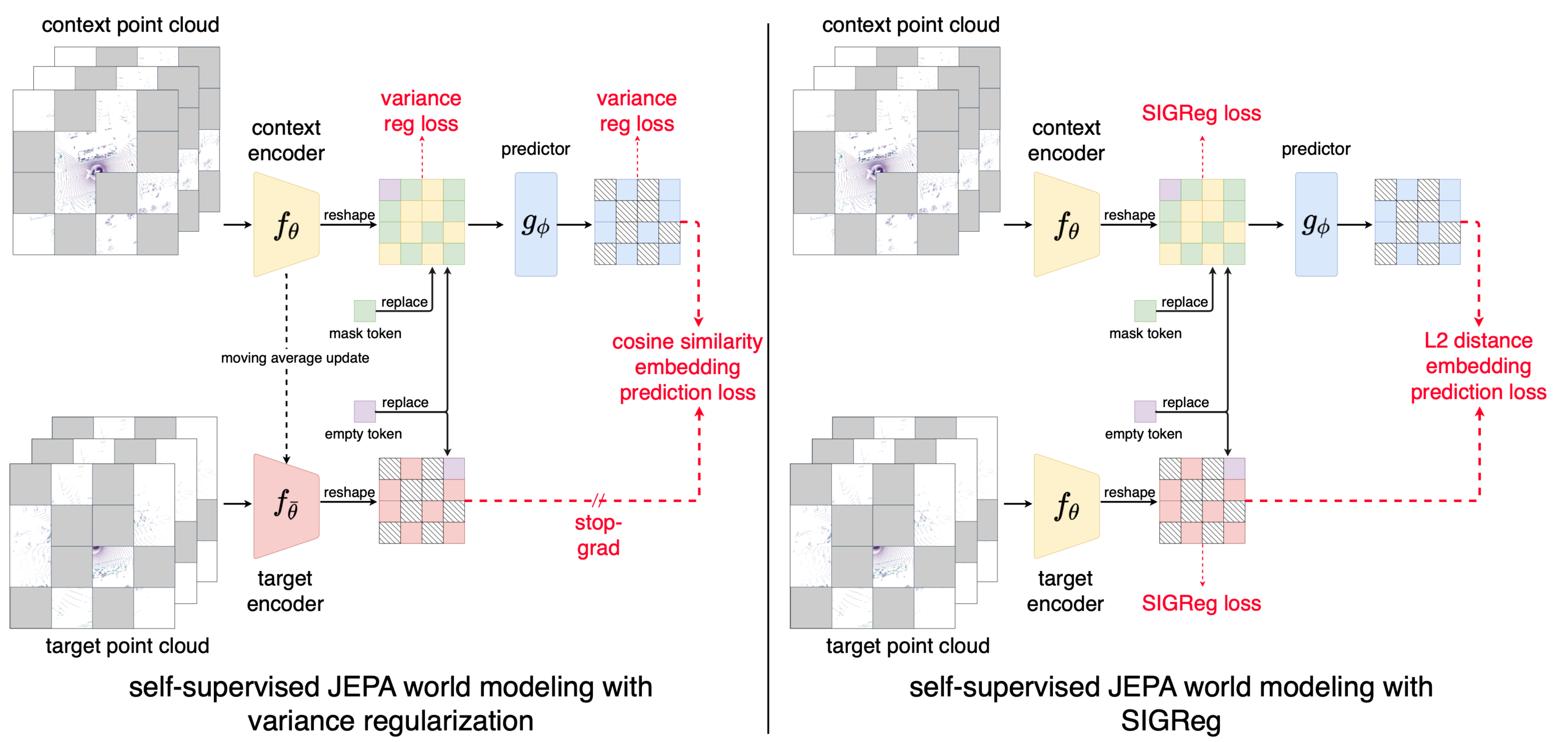}
\caption{Overview of the Phase 1 framework with variance regularization (left) and SIGReg regularization (right) variants (best viewed when zoomed in).}
\end{center}
\label{fig:architecture}
\vspace{-0.3in}
\end{figure}

The loss consists of an embedding prediction loss and a regularization loss at non-empty grids:
\begin{equation}
\mathcal{L} = \mathcal{L}_{\text{jepa}} + \lambda_{\text{reg}} \mathcal{L}_{\text{reg}},
\end{equation}
Specifically, mask tokens and empty tokens are used to represent masked regions and empty-region embeddings, respectively, which is the same as that used in AD-L-JEPA~\cite{zhu2025ad} The pretraining objective includes a cosine-similarity--based embedding prediction loss applied to masked grids, along with a variance regularization loss applied to non-empty grids at the outputs of both the encoder and the predictor. We also experimented with SIGReg~\cite{balestriero2025lejepa} as the regularization loss. In this setting, no moving-average update is required: the context encoder and the target encoder are identical. SIGReg is applied to the outputs of both encoders, and the prediction loss is the L2 distance rather than cosine similarity. An overview of the network architecture variants during phase 1 is shown in Figure~\ref{fig:architecture}.

\subsection{Phase 2: Evaluating World Models with OCF finetuning}
We evaluate world model representations on a downstream OCF task that captures dense scene geometry across frames. A sparse 3D encoder pretrained in Phase~1 extracts per-frame BEV features, which are concatenated into multi-frame representations. To assess the JEPA encoder’s representation quality, we keep the decoder as simple as possible by using a lightweight decoder with three convolutional layers to predict future BEV features, followed by a linear layer for per-grid binary classification and interpolation to the target OCF resolution. The model is trained with binary cross-entropy loss.

\vspace{-0.15in}
\section{Experiments}
\vspace{-0.15in}
We use Waymo dataset~\cite{sun2020scalability} for all experiments with OpenPCDet framework for LiDAR. Details on obtaining the training labels are provided in the supplementary materials as an extended explanation of the original paper~\cite{liu2024lidar}. We conduct Phase~1 pretraining for 30 epochs using 8 A100 GPUs with the Adam optimizer and a one-cycle learning rate scheduler, with a learning rate of 0.0003 and a weight decay of 0.01. For the variance regularization setting, we use a total batch size of 32, with the other hyperparameters set to the default values of AD-L-JEPA~\cite{zhu2025ad}. For SIGReg, we train for 30 epochs using 8 A100 GPUs with a total batch size of 16; $\lambda_{\text{reg}}=10$ for variance regularization and $\lambda_{\text{reg}=0.001}$ for SIGReg. The masking ratio is set to 50\% for non-empty grids and 50\% for empty grids. 
For Phase~2 fine-tuning, we use a single A100 GPU with a batch size of 4 and a constant learning rate of 0.0005, and fine-tune the model for 3 epochs to predict the next 5 frames' OCF label given the past 5 frames' sparse input. For each fine-tuning run, we use three random seeds and report the mean and standard deviation in Table~\ref{tab:exp}. The results show that the pretrained encoder of AD-LiST-JEPA outperforms training from scratch. Furthermore, the recently proposed SIGReg~\citeyear{balestriero2025lejepa} significantly outperforms variance regularization loss for this simple network architecture designed for the downstream task, indicating a promising direction for purely regularization-based methods to avoid representation collapse.

\begin{table*}[h]
\label{tab:exp}
\centering
\begin{tabular}{l| c |  c}
\hline
Method & IoU$_{\text{full}}$(\%) & IoU$_{\text{close}}$ (\%)  
\\
\hline

scratch, linear   & 38.56 $\pm$ 0.19 & 42.87 $\pm$ 0.17\\
\hline

AD-LiST-JEPA, small   & 39.09 $\pm$ 0.36 & 43.43 $\pm$ 0.39\\
\hline
AD-LiST-JEPA, small, SIGReg   &  \textbf{39.35} $\pm$ 0.24 & \textbf{43.70} $\pm$ 0.24\\
\hline
AD-LiST-JEPA, full   &  39.01 $\pm$ 0.47& 43.46 $\pm$ 0.44\\
\hline
AD-LiST-JEPA, full, SIGReg   &  \textbf{39.41} $\pm$ 0.31 & \textbf{43.86} $\pm$ 0.30\\
\hline
\end{tabular}
\caption{Occupancy completion and forecasting segmentation performance on the Waymo dataset. ``Small'' and ``large'' denote Phase~1 pretraining with 190 and 950 Waymo sequences, respectively. IoU$_{\text{full}}$ (\%) is computed over the full range, while IoU$_{\text{close}}$ is computed over a closer region with half the spatial extent in both $x$ and $y$.}
\end{table*}

\vspace{-0.25in}
\section{Conclusion}
\vspace{-0.15in}
This paper presents a proof of concept for training JEPA-based spatiotemporal world models on multi-frame LiDAR data for autonomous driving. We evaluate the quality of the learned world models using a downstream occupancy completion and forecasting task. Our results indicate promising directions for future research; experiments on larger datasets and with larger network architectures will be explored in future work.

\newpage
\clearpage

\bibliography{iclr2026_conference}
\bibliographystyle{iclr2026_conference}

\newpage
\clearpage

\appendix
\section{Appendix}
\begin{figure}[ht!]
\begin{center}
\includegraphics[width=\linewidth]{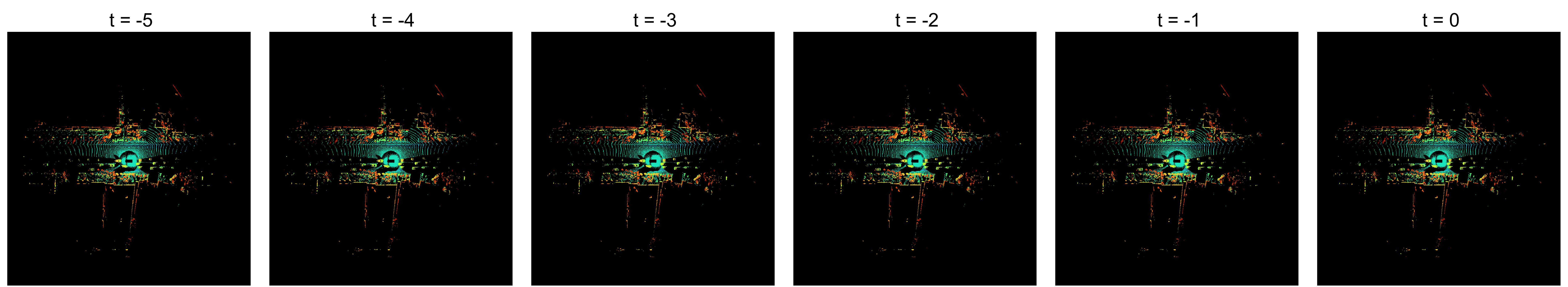}
\vspace{-0.25in}
\caption{Raw past (including currrent) point cloud  visualization as the neural network input from time -5 to 0. The color represents height relative to the ground.}
\end{center}
\label{fig:ocf_input}
\end{figure}

\begin{figure}[ht!]
\begin{center}
\includegraphics[width=\linewidth]{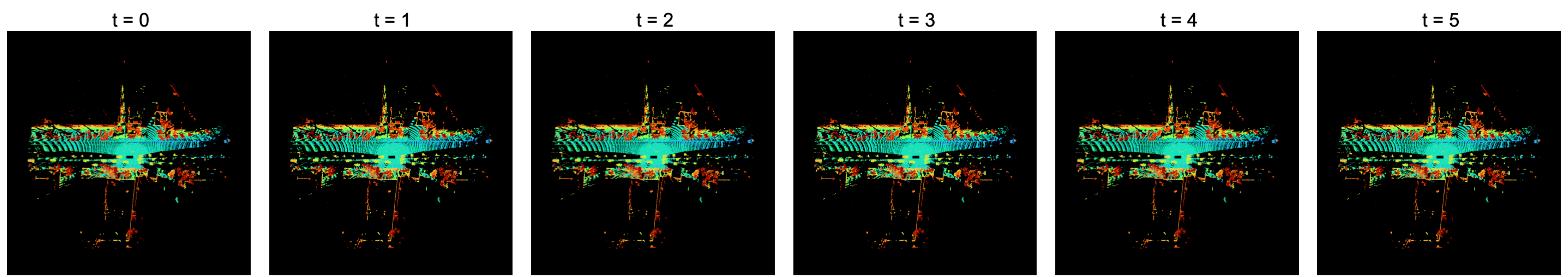}
\vspace{-0.25in}
\caption{Processed future (including current) completed occupancy visualization as the neural network prediction target from time 0 to 5. The color represents height relative to the ground.}
\end{center}
\label{fig:ocf_label}
\end{figure}

This part is not novel; it is an extended description of the data processing pipeline in~\cite{liu2024lidar}, with pseudocode to help readers better understand the data processing pipeline. The visualization of past raw input point clouds is shown in Figure~\ref{fig:ocf_input}, while the future completed occupancy ground-truth labels after data processing are visualized in Figure~\ref{fig:ocf_label}.

The data processing contains two phase: 
\begin{itemize}
    \item LiDAR sequence transformation
    \item voxelization with ray casting
\end{itemize} 

\subsection{LiDAR Sequence Transformation}
The ultimate goal of this component is to aggregate points belonging to the same object instance across previous and subsequent frames. First, a coordinate transformation is applied to map ego-centric points into a unified reference coordinate system at the current time step. This process includes ghost object removal, which eliminates objects that appear in previous or subsequent frames but are absent in the current frame, as well as moving object alignment for objects present in the current frame. Points from these objects are then aggregated using instance bounding boxes from other frames. Singular Value Decomposition (SVD) is used to solve the orthogonal Procrustes problem for this aggregation~\cite{kabsch1976solution}. The algorithm is summarized in Algorithm~\ref{algo:ocf_process_1}.

\begin{algorithm}
\caption{LiDAR Sequence Transformation}
\begin{algorithmic}[1]  
\Procedure{Transform}{$\text{id}_\text{cur}$}
    \State $L \gets \text{total length of the sequence}$    
    \State $p_\text{dict}^{\text{cur}} \gets \{\}$ 
    \For{each frame $i \gets \text{id}_\text{start} \ \textbf{to} \ \text{id}_\text{end} \   \textbf{and} \ i \neq \text{id}_\text{cur}$}
        \State $\text{pose}_i \in \mathbb{R}^{4 \times 4}\gets \text{pose at time index i}$

        \State $B_{\text{cur}} \gets \text{instance boxes at time } \text{id}_{\text{cur}}$
        \State $B_i \gets \text{instance  boxes at time i}$ 
        \For{ each instance box $j \gets 0$ \textbf{to} $|B_i|-1$}
            \If{$B_{i}{[j]} \notin B_{\text{cur}}$}
                \State delete points for instance $j$ at time $i$
            \ElsIf{{$B_{i}{[j]} \in B_{\text{cur}}$}}
               
                \State $m \gets \text{index of the same instance in } B_{\text{cur}}$
                \State $(R, c) \gets \operatorname{SVD\_Align}(B_i[j], B_{\text{cur}}[m])$
                \State $p_j \gets \text{points of in } B_{i}{[j]}$
                \State $T_\text{cur}^\text{global} \gets \text{  current to global pose}$
                \State $T_i^{\text{global}} \gets \text{time i to global pose}$
                \State $\tilde{p}_j = {\{T_i^{\text{global}}\}}^{-1} T_\text{cur}^\text{global} \begin{bmatrix} R^{\mathsf{T}} p_j + c \\ \mathbf{1} \end{bmatrix}$
                \State replace $p_j$ with $\tilde{p}_j$ for points at frame i      
            \EndIf
        \EndFor
        \State $p_\text{dict}^{\text{cur}}[i] \gets \text{updated points at frame i}$
    \EndFor
    \State \Return $p_\text{dict}^{\text{cur}}$
\EndProcedure
\end{algorithmic}
\label{algo:ocf_process_1}
\end{algorithm}

\begin{algorithm}
\caption{Obtain Network Input and Outputs}
\begin{algorithmic}[1]  
\Procedure{Transform}{$\text{id}_\text{cur}$}
    \State $L \gets \text{total length of the sequence}$    
    \State $R \gets (x_\text{min}, y_\text{min}, z_\text{min}, x_\text{max}, y_\text{max}, z_\text{max})$,  \text{range} 
    \State $V \gets (v_x^\text{size}, v_x^\text{size}, v_x^\text{size})$, \text{voxel size}

    \State $\tilde{O}^{\text{sparse\_iput}}=[]$
    \For{each frame $i \gets \text{id}_\text{cur}-N_\text{pre}  \ \textbf{to} \ \text{id}_\text{cur}$ }   
        \State $P_i^{all} \gets \text{all points at time i}$
        \State $T_i^{\text{global}} \gets \text{time i to global pose}$
        \State $\tilde{P}_i^{all} \gets {\{T_{\text{cur}}^{\text{global}}\}}^{-1} T_i^{\text{global}} P_i^{all}$
        \State $q_i^\text{origin} \gets \text{sensor origin at time i}$
        \State $O_i^{\text{sparse\_occ}}, \_ \gets 
        \operatorname{RayCast(q_i, \tilde{P}_i^{all}, R, V)}$
        \State $\tilde{O}^{\text{sparse\_iput}}.\operatorname{append(\tilde{P}_i^{all})}$
        
    \EndFor

    \\
    
    \State $O^{\text{occ}} =[]$
    \State $O^{\text{invalid}}$
    \For{each frame $i \gets \text{id}_\text{cur} \ \textbf{to} \ \text{id}_\text{cur}+N_\text{post}$ }   
        \State $T_i^{\text{global}} \gets \text{time i to global pose}$
        \State $K_\text{start} \gets \operatorname{max(i-\text{N}_\text{pre}, 0)}$
        \State $K_\text{end} \gets \operatorname{min(i+\text{N}_\text{post}, \text{L})}$
        \State $P_i^{\text{aggr}}\gets []$
        \For{each frame $j \gets K_\text{start} \ \textbf{to} \ K_\text{end}$}
            \State $P_j^{all} \gets \text{all points at time j}$
            \State $T_j^{\text{global}} \gets \text{time j to global pose}$
            \State $\tilde{P}_j^{all} \gets {\{T_\text{cur}^{\text{global}}\}}^{-1} T_j^{\text{global}} P_j^{all}$ 
            \State $P_i^{\text{aggr}}.\operatorname{append(\tilde{P}_j^{all})}$
        \EndFor
        \State $q_i^\text{origin} \gets \text{sensor origin at time i}$
        \State $O_i^{\text{occ}}, O_i^{\text{invalid}} \gets \operatorname{RayCast(q_i^\text{origin}, P_i^{\text{aggr}}, R, V)}$
        \State $O^{\text{occ}}.\operatorname{append(O_i^{\text{occ}})}$
        \State $O^{\text{invalid}}.\operatorname{append(O_i^{\text{invalid}})}$
        
    \EndFor
    \State \Return $(\tilde{O}^{\text{sparse\_iput}}, O^{\text{occ}}, O^{\text{invalid}})$
\EndProcedure
\end{algorithmic}
\label{algo:ocf_process_2}
\end{algorithm}

\subsection{Voxelization with Ray Casting}
After Algorithm~\ref{algo:ocf_process_1}, a densified point cloud is obtained by aggregating instances across time. To generate the completed occupancy, voxelization with ray casting is then applied. Rays are cast from the origin to each point to determine whether voxels along each ray are occupied (contain points), empty (contain no points), or invalid (no rays pass through them). The resulting occupancy and invalid masks are used for network training and evaluation. The algorithm is summarized in Algorithm~\ref{algo:ocf_process_2}.

\end{document}